\documentclass{article} 

\usepackage{geometry} 
\usepackage{amsmath}  
\usepackage{graphicx}  
\usepackage{algorithm2e}
\usepackage{amsfonts} 
\usepackage{graphicx}
\usepackage{makecell}
\usepackage{verbatim}
\usepackage{mathtools}
\usepackage{subcaption}
\usepackage[nottoc]{tocbibind}
\title{How much data do you need? Part 2: Predicting DL class specific training dataset sizes.}
\author{Thomas Mühlenstädt, Jelena Frtunikj}

\graphicspath{ {./plots/} }
\RestyleAlgo{ruled}

\newcommand{\nosemic}{\SetEndCharOfAlgoLine{\relax}}
\newcommand{\dosemic}{\SetEndCharOfAlgoLine{\string;}}
\newcommand{\pushline}{\Indp}
\begin{document}
\maketitle

\section*{Abstract}

This paper targets the question of predicting machine learning classification model performance, when taking into account the number of training examples per class and not just the overall number of training examples.
This leads to the a combinatorial question, which combinations of number of training examples per class should be considered, given a fixed overall training dataset size. In order to solve this question, an algorithm is suggested which is motivated from special cases of space filling design of experiments.
The resulting data are modeled using models like powerlaw curves and similar models, extended like generalized linear models i.e. by replacing the overall training dataset size by a parametrized linear combination of the number of  training examples per label class. 
The proposed algorithm has been applied on the CIFAR10 and the EMNIST datasets.

\section{Introduction and relevant related literature}

The main motivation for this research comes from the question of how much more data needs to be labeled to achieved in order to gain a certain machine learning model performance, for example looking at accuracies or $f_1$ scores. There exist many methods in literature dealing with this question, which will be reviewed in the next paragraph. However, just knowing how many additional samples are needed does not answer how these samples should be distributed across different classes. In order to address this question, methods are developed to have a more detailed look at how different distributions among the label classes in a classification dataset are influencing model performance.

This section presents the related literature from different domains, e.g. \cite{hestness2017deep} presents empirical characterization of the generalization error and model size growth as training sets grow.  The empirical results show power-law generalization error scaling, resulting in power-law exponents that need to be determined empirically and which influence the "steepness" of the learning curve. Prior work theoretically predict that the scaling exponent $\beta$ is in the range of $-1$ and $-0.5$, however the authors of the paper and the empirical studies performed on real applications show that $\beta$ usually settles between $-0.35$ and $-0.07$. The authors assume that this behaviour is due to fact that the scaling law is dependent on aspects of the problem domain or data distribution. The also authors show that the learning curve begins in the small data region, where models struggle to learn from a small number of training samples i.e. the model performs as “random" guessing. The middle portion of learning curves is the power-law region, where each new training sample provides information that helps models improve predictions on previously unseen samples. The end portion of the curve is called the “irreducible error region” i.e. lower-bound error past which models will be unable to improve. The reasons for such “irreducible” behaviour are e.g. mislabeled samples in the training or validation data sets. The authors have performed empirical studies on four machine learning domains: machine translation, language modeling, image processing, and speech recognition.

The authors of \cite{mahmood2022data} go beyond estimation of data set requirements from power law function and investigate three alternative regression functions i.e. Arctan, Logarithmic, Algebraic Root. They show that all of the functions are well-suited towards estimating model performance however each function is almost always either overly optimistic i.e. under-estimating the data requirement or pessimistic i.e. over-estimating. This means that there is no one best regression function for all situations. Through simulation of a data collection workflow the paper shows that incrementally collecting data over multiple rounds and combining those with techniques that under-estimate leads to collecting up to 90\% of the true amount of data needed. In addition the authors introduce a correction factor that can be learned by simulating on prior tasks and which helps to better fitting functions. The fitting of the three regression functions and the benefits of the usage of the correction factor have been applied on classification, detection, and segmentation tasks with different data sets incl. CIFAR10.

The paper \cite{cho2016data} focuses on predicting the needed training dataset size for achieving necessary accuracy for a classification task in the medical domain. More precisely the use case is classifying axial Computed Tomography (CT) images into six anatomical classes. The authors apply the already introduced (\cite{Figueroa2012PredictingSampleSizeForClass}) power law fitting to the mentioned use case. This paper focuses only on the medical domain and provides among the first empirical results that the power law fitting for determining the training data set size also applies for the medical domain.

Inspired by the fact that power law scaling of the error w.r.t. data suggests that many training examples are redundant, the authors of \cite{sorscher2023neural} investigate the hypothesis of pruning training datasets to much smaller sizes and training on the smaller pruned datasets without sacrificing performance. The authors show in theory and in practice that one can break beyond power law scaling and even reduce it to exponential scaling if one uses an efficient data pruning metric that ranks the order in which training examples are being discarded. The theory proof reveals two very interesting points: 1) the optimal pruning strategy changes depending on the amount of initial data i.e. with abundant (scarce) initial data, one should retain only hard (easy) examples and 2) exponential scaling is possible w.r.t. pruned dataset size only if one chooses an increasing Pareto optimal pruning fraction as a function of initial dataset size. The authors demonstrate empirically the exponential scaling of the error w.r.t. pruned dataset size for a ResNets trained from scratch on SVHN, CIFAR-10 and ImageNet, and Vision Transformers fine-tuned on CIFAR-10. In addition, the paper presents a new and comparably good unsupervised data pruning metric that does not require labels, unlike other prior unsupervised pruning metrics that require labels and much more compute. The k-means based pruning metric (clustering in the embedding space of an ImageNet pre-trained model) enables discarding 20\% of ImageNet data without sacrificing performance, on par with the best and most compute intensive supervised metric.

Motivated by the fact that GPUs and model capacity have continuously been growing, but training datasets have remained stagnant, the authors of \cite{sun2017revisiting} investigate the impact of big datasets (i.e. the JFT-300M) on the performance of vision tasks (e.g. object detection, segmentation etc. ). The JFT dataset contains more than 300M images that are labeled with 18291 annotations and the authors use the ResNet-101 model architecture in the experiments. The experiments cover that performance increases logarithmically based on the volume of training data. This is valid for both use cases i.e. when using fine-tuning and when using freezing feature extractors. The authors perform also further experiments in which they show couple of more interesting findings e.g. 1) capacity is crucial i.e. in order to fully use the 300M images, one needs higher capacity models. For example, in case of ResNet-50 the gain on COCO object detection is much smaller (1.87\%) compared to (3\%) when using ResNet-152. 2) JFT-300M has 18K labels in total and in order to understand what the large number of classes brings, the authors select a subset of 941 labels which have direct correspondence to the 1000 ImageNet labels. The experiments show that performance benefit comes from more training images instead of more labels.

Compared to the other papers that mainly target vision tasks, \cite{hoffmann2022training} focus on large language model. The trend so far in large language model training has been to increase the model size, often without increasing the number of training tokens. Thus the paper investigates the optimal model size and number of tokens for training a transformer language model under a given compute/FLOPs budget. The paper presents three predictive approaches that start by training a range of models varying both model size and the number of training tokens and use the resulting training curves to fit an empirical estimator. Similarly as some of the above presented literature the paper assumes a power-law relationship between compute and model size. The resulting predictions are similar for all three methods and suggest that parameter count and number of training tokens should be increased equally with more compute i.e. for every doubling of model size the number of training tokens should also be doubled. The authors propose and use the Chinchilla LLM to test and proof the hypothesis.

The authors of \cite{alabdulmohsin2022revisiting} revisit the scaling law in vision and language domain and emphasise that in order to achieve benefits of such scaling laws in practice, it is important that scaling laws extrapolate accurately instead of interpolating the learning curve. The paper shows how the scaling exponent c of a power law estimator that extrapolates best can be quite different from the exponent that best fits the given (finite) learning curve. Thus a new estimator (denoted as M4) is proposed which extrapolates more accurately than previous methods such as a power law estimator M2. M2 is a subset of M4 such that using M4 becomes equivalent to using M2 when the observed learning curve resides entirely in the power law regime. Moreover, the M4 estimator has a sigmoid shape and that contributes to the better performance. The new proposed estimator M4 has been validated in several domains such as image classification, neural machine translation, language modelling, and other related tasks. For the validation the authors propose a more rigorous validation of scaling law parameters based on the extrapolation loss.

Compared to the previous mentioned papers where the scaling law has been empirically determined for a diverse of problem types (supervised, unsupervised, transfer learning) and data types (images, video, text) and many NN architectures (Transformers, ConvNets, ...), \cite{hutter2021learning} focuses on the theoretical understanding of the phenomena (error scaling w.r.t. data size n). In finite-dimensional models, the error typically decreases with n - 1/2 or n - 1 , where n is the sample size. The author of the paper also argues that noise rapidly improves in time-averaged learning curves for increasing n, which means that the model selection should be based on a cumulative AUC or time-averaged error and not final test error which is often done in practice.



\section{Method overview}

The task considered in this paper is image classification, i.e. we aim to find a model $y = f_{\theta}(x)$, where $x$ is an image tensor and $y$ being a one hot encoding for the classes contained in the dataset.
The training data consists of $k$ classes, with each class $j$ having $n_j^{max}$ labeled images in the training dataset with $n^{max} = \sum_{j= 1}^k n_j^{max}$ being the overall number of training images.
For completeness, we assume there is a labelled test and/or validation data set used to calculate the performance of a trained model $f_{\hat{\theta}}(x)$.
The performance metric used is accuracy, denoted as $a(f_{\hat{\theta}}(.))$. In general, the methods described here would also for other metrics like e.g. $f_1$ scores, precision or recall.
We refer to standard literature like e.g. \cite{Goodfellowetal2016} and \cite{Bishop2006PatternRecog} for these general details.
As outlined before, most of the literature for neural scaling laws focusses on assuming different training dataset sizes $n^{train}$ for a number of different training jobs, i.e. to find a function $g_{\omega}(n^{train})$ which predicts the chosen performance metric for different training dataset sizes.
This means implicitly that each class has the same importance for the performance.
However, this is not necessarily true, some classes might be more difficult to train on than others.
Hence here an approach is taken, which allows to estimate the importance of training images from each class individually.
Also, the compute budget is an important aspect of performance, hence we also aim at considering the different number of training epochs affects performance.
Thus we expand the function $g_{\omega}(n^{train})$ to be more general: $g_{\omega}(n_1^{train}, \dots, n_k^{train})$ shall be a function predicting the performance of our model depending on the individual class training image counts.
Even more general, we can also incorporate the number of training epochs $n^{epoch}$ used for estimating the parameters $\theta$ into the function, now becoming $g_{\omega}(n_1^{train}, \dots, n_k^{train}, n^{epoch})$.
In order to fit functions which allow for estimation of more differentiating effects between classes and epochs, more divers training datasets need to be generated.
Hence one focus of this paper is to suggest an algorithm which allows generation of divers training data sets.
For modeling these data sets and making predictions and importance statements, non-linear models like powerlaw curves are fitted in a way that the input is a linear combination of the training dataset size per class plus the number of epochs trained.
The coefficients of this linear combination are the parameters to be estimated.

\subsection{Data generation algorithm}



For a given training dataset, a smaller subset can be sampled randomly by assigning the same probability of being included in the sample to each image.
Although in general this is a good sampling strategy in many contexts, here it is not what is desired.
Especially for large training datasets, the law of large numbers says that the class proportions in the sample subset will be very similar to the class distribution in the complete training dataset.
In order to achieve a number of divers training datasets, both with respect to the total size as well with respect to the class proportions, an algorithm is suggested in the following.
It takes motivation from a special case of statistical design of experiments, notably constrained space filling mixture designs. Please see \cite{gomes_hal_spacefilling_mixtures} for more information on spacefilling mixture desigsn.
Design of experiments is a set of methods, originating back to as early as Sir Ronald Fishers book on the topic \cite{fisher_1935}, which all aim at collecting optimal data for fitting a specific statistical/mathematical model. This might be simple linear models or much more complex
models like nonlinear, multidimensional models \cite{myers2009response}, \cite{Goos_Jones_optDoE_2011}. In contrast to many machine learning and active learning methods, design of experiments
often aims at finding the minimal dataset with which a model can be fitted.
Mixture designs in general are a class of designs originating out of chemistry: The composition of some ingredients to a chemical experiments needs to sum up to 100\%, see for example \cite{Nist_2012_eng_stats}, chapter 5.5.4 for a short introduction.
The reason why this is fitting here is that, given a target dataset size $n_{subset}$, we want to create a number of different combinations of training datasets, which all have the same total number of training images but distributed differently among to the different classes.
There are different ways to describe the optimality of a design of experiments. Here we choose an optimality criterion which is considered space filling, the so called maximin optimality criterion, which tries to maximize the minimum (euclidean) distance between any two data points in the design.
The reasoning behind this is that having a data point $d_2 \in D$ , which are very close to a data point $d_1 \in D$ does not bring additional benefit, as we have learned already the response of
our target function $g(.)$ at position $d_1$.
This choice is to a big degree subjective. We have chosen this optimality criterion here, as we want to generate a wide spread of different combinations of class counts, given a fixed overall count of training images.
The design needs to be constrained, as for larger subset sizes, it might happen that a purely randomly generated combination of class counts is for some classes higher than the maximum number of available training images per class.
The algorithm (Algorithm 1) to find an optimal design has 2 stages. First, in the initialization phase, a design is constructed which is fullfilling the following: 
each experiment row is summing up to the target dataset size and the constraint for the maximum number of images per class is fullfilled.
In the second stage, this design is improved iteratively by a pointwise exchange algorithm: The pair of design points with the minimal distance is found and one of them is replaced by a new, randomly generated candidate design point.
If the maximin criterion for the new candidate design is improved compared to the previously best design, the candidate design is accepted as new best design.
Otherwise, the candidate design is rejected. This process is repeated a fixed number of times.
Similar pointwise optimization procedures are quite common in the area of design experiments and are rooting back to early references like \cite{fedorov1972theory}.
One important detail in this process is how design points are suggested, during the initialization phase as well as during the optimization phase.
For the unconstrained case, candidate points can be suggested by using standard uniform random sampling. In the case of heavily constrained mixture, a heuristic can be applied to increase the chance to sample a candidate fullfilling the constraint. For details, please check the appendix.

\begin{algorithm}
    \SetKwInOut{Input}{input}\SetKwInOut{Initialize}{initialize}\SetKwInOut{Optimize}{optimize}
    \caption{Mixture Design creation} \label{alg:one}
    \nosemic
    \Input{\;
        \dosemic
        \pushline
        Define $n_{doe} \in \mathbb{N}$: The number of experiment rows to be generated\;
        Define $n_{cmax} \in \mathbb{N}$: The maximum number of training images in each class\;
        Define $n_{opt} \in \mathbb{N}$: The maximum number of iterations used for optimizing the DoE\;
        Define $n_{candidate} \in \mathbb{N}$: The batch size for generating candidate mixture data points\;
    }
    \Initialize{\;
        \dosemic
        Intialize $\tilde{doe}$ as empty dataframe\;
        Set $n_{doe} = 0$\;
        \While{$\tilde{n}_{doe} < n_{doe}$}{
            Create a batch of candidate mixture data points $candidates$ of size $n_{candidate}$\;
            For each candidate point: Delete candidate point, if for any class $c$ $n^{c} > n_{cmax}$\;
            Add all remaining rows of $candidates$ to $\tilde{doe}$\;
            set $n_{doe} = $ number of rows of $doe$ \;
        }
        Set $doe$ to the first $n_{doe}$ rows of $\tilde{doe}$\;
    }
    \Optimize{\;
        \dosemic
        Set $doe_{opt} = doe$\;
        Set $Mm = \min_{t \neq b}(\|x_t - x_b\|)$ \;
        \For{$i \leq n_{opt}$}{
            Set $\tilde{doe} = doe_{opt}$\;
            Determine point pair $k \neq j$ with $\|x_k - x_j\| = \min_{x_t \neq x_b \in \tilde{doe}}(\|x_t - x_b\|) $\;
            Create a candidate point $\tilde{x}$ with $max(x^{c}) \leq n_{cmax}$ (fulfilling class constraints)\;
            In $\tilde{doe}$, replace point $x_k$ with $\tilde{x}$\;
            \If{$\min_{x_t \neq x_b \in \tilde{doe}}(\|x_t - x_b\|) > Mm$}{
                Set $doe_{opt} = \tilde{doe}$\;
                Set $Mm = \min_{x_t \neq x_b \in \tilde{doe}}(\|x_t - x_b\|)$\;
            }
        }
    }
\end{algorithm}

The following algorithm shows the complete experiment workflow which uses the above defined mixture design to create the datasets used in the experiments. These experiments generate the needed data points used for the model fitting.

\begin{algorithm}
    \SetKwInOut{Input}{input}
    \caption{Data generation}\label{alg:two}
    \nosemic
    \Input{ \;
        \dosemic
        \pushline
        Choose an image classification dataset\;
        Choose a network architecture $f_{\theta}(x)$ suitable for the chosen image classification\;
        Define a sequence $subsets$ of growing integers between $0$ and $n_{train-max}$\;
        Define $n_{repeat} \in \mathbb{N}$: The number of rows to be generated per $s \in subsets$\;
        Define $e_{max} \in \mathbb{N}$: The maximum number of epochs to train a network \;
        Define a sequence $e_{check}$ with elements $c \in \mathbb{N}, c \le e_{max}$: The epochs where to evaluate performance of
    }
    \For{$s \in subsets$}{
        Generate mixture design $D_s$ for subset size $s$ with $n_{repeat}$ rows\;
        \For{$i \in (0, \dots,  n_{repeat} - 1)$}{
            Generate training dataset $T_{i, s}$according to row $D_{s, i}$\;
            Initialize network $f_{\theta}$\;
            \For{$e \in (0, \dots, e_{max})$}{
                Train network $f_{\theta}$ for one epoch\;
                \If{$e \in check_e$}{
                    Evaluate the performance of $f_{\theta}$ on the test dataset\;
                    Write the evaluation results (losses, metrics) to a logger\;
                }
            }

        }
    }
\end{algorithm}

\subsection{Scaling law model fitting}

So far the standard way of fitting a model to the data is a non-linear least squares fit with different types of underlying model equations, all depending on the overall training dataset size $n$.
Non-linear least squares are implemented in the scipy function \verb|curve_fit| (\cite{2020SciPy}).
The power law scaling (\cite{mahmood2022data}) is a frequently chosen option: $g_{\omega}(n) = \omega_1 n^{\omega_2} + \omega_3$.
But also arctan scaling ($g_{\omega}(n) = \frac{200}{\pi} \arctan(\omega_1 \frac{\pi}{2}n + \omega_2) + \omega_3$),
logarithmic scaling ($g_{\omega}(n) = \omega_1 \log(n + \omega_2) + \omega_3$) and
algebraic root ($g_{\omega}(n) = \frac{100n}{1 + \| \omega_1 n\|^{1/\omega_2}} + \omega_3$) are possible choices.
For each of these models, we can replace $n$ by a linear combination of $n_c$ with a parameter $\beta_c$: $\sum_{c = 1}^C \beta_c n_c$.
Another modification which is done here is to use a combination of different modelling functions, specifically a combination of the powerlaw model and an arctan model.
Also, as the modifications introduced above might result in a high number of model effects for datasets with many classes, forward selection as a feature selection method is applied in one of the examples to only include features which actually have a predictive power.

\section{Experiments using CIFAR10}

\subsection{Experiment setup}

\begin{figure}
    \begin{subfigure}{.5\textwidth}
        \centering
        \includegraphics[width=.8\linewidth]{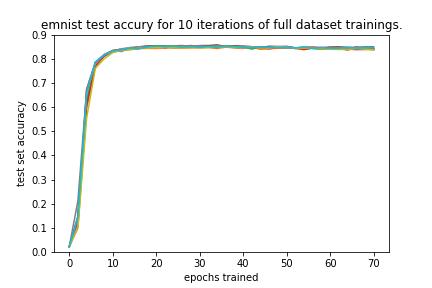}
        \caption{Full training results for 10 separate training runs (EMNIST dataset).}
        \label{fig_full_dataset_epoch_vs_acc}
    \end{subfigure}%
    \begin{subfigure}{.5\textwidth}
        \centering
        \includegraphics[width=.8\linewidth]{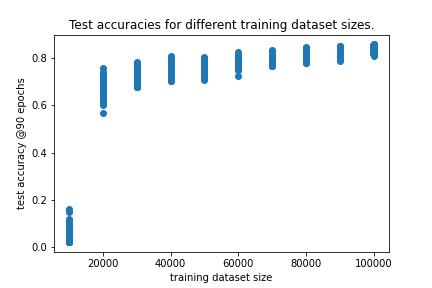}
        \caption{Test accuracies vs. training dataset sizes (EMNIST dataset).}
        \label{fig_traing_subset_size_vs_test_acc}
    \end{subfigure}

    \caption{Descriptive results for the models fitted to EMNIST dataset.}
    \label{fig:desc_plots_emnist}
\end{figure}

The CIFAR10 dataset \cite{Cifar10} is a well known benchmark dataset for image classification.
It consists of 50000 training images, in 10 classes, each class having 5000 training images of the shape 32x32x3 and 10000 test images of the same shape, also equally distributed across the 10 classes.
A standard Resnet18 architecture \cite{he2015resnet} is used for classification as standard model as implemented in the pytorch \cite{pytorch} model zoo.
For optimizing this model, a standard SGD optimizer with cross entropy loss is used with \verb|learning_rate=0.1, momentum=0.9| and \verb|weight_decay=1e-4|, together with a learning rate scheduler, reducing the the learning at 100 (which is also clearly visible at severals accuracy vs. epoch plots, e.g. Figure \ref{fig_full_dataset_epoch_vs_acc}) and 150 epochs.
For data preparation/transformations, random horizontel flipping, random cropping and batch normalization with predefined mean and standard deviation is used.
With these settings, validation dataset accuracies up to 84\% are achieved.
There are for sure more advanced model architectures for the CIFAR10 dataset, however we aimed at balancing an at least medium performant model with a having a reasonable training time, as the training process is repeated here a high number of times.
For the experiments done here, we applied the algorithm from section 4 with the following settings:
As subset sizes we have chosen $[5000, 10000, \dots, 40000, 45000]$. For each subset size a design of experiments according to Algorithm \ref{alg:two} with 30 different settings has been created for training a function $g_{\omega}$ as well as a validation dataset with 15 runs per subset size.
For each training dataset, the same model and hyperparameter settings has been used, training for up to 195 epochs, checking for validation accuracy every 5 epochs.

\begin{figure}
    \begin{subfigure}{.5\textwidth}
        \centering
        \includegraphics[width=.8\linewidth]{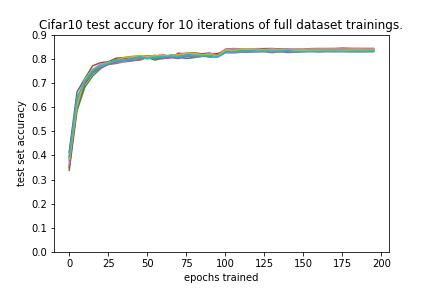}
        \caption{Full training results for 10 seperate training runs.}
        \label{fig_full_dataset_epoch_vs_acc_cifar}
    \end{subfigure}%
    \begin{subfigure}{.5\textwidth}
        \centering
        \includegraphics[width=.8\linewidth]{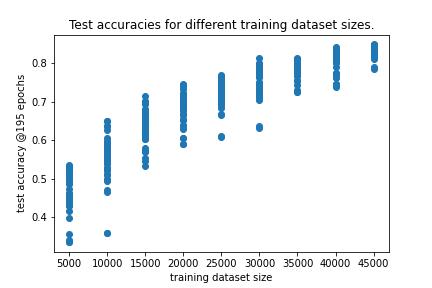}
        \caption{Test accuracies vs. training dataset sizes.}
        \label{fig_traing_subset_size_vs_test_acc_cifar}
    \end{subfigure}

    \caption{Descriptive results for the models fitted to CIFAR10 dataset.}
    \label{fig:desc_plots_cifar10}
\end{figure}

\begin{figure}
    \begin{subfigure}{.5\textwidth}
        \centering
        \includegraphics[width=.8\linewidth]{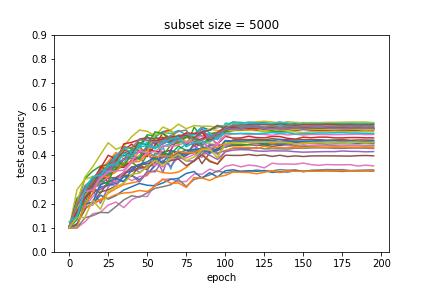}
        \caption{training dataset size = $5000$ images}
        \label{fig:subsetsize5000}
    \end{subfigure}%
    \begin{subfigure}{.5\textwidth}
        \centering
        \includegraphics[width=.8\linewidth]{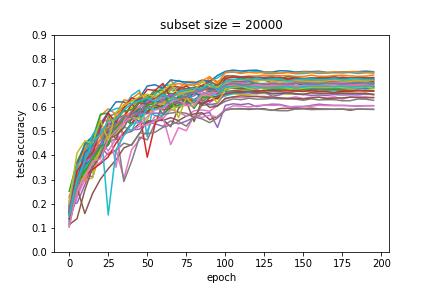}
        \caption{training dataset size = $20000$ images}
        \label{fig:subsetsize20000}
    \end{subfigure}
    \begin{subfigure}{.5\textwidth}
        \centering
        \includegraphics[width=.8\linewidth]{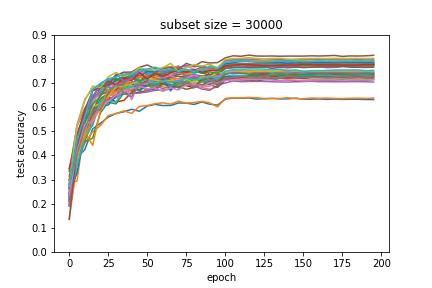}
        \caption{training dataset size = $30000$ images}
        \label{fig:subsetsize30000}
    \end{subfigure}%
    \begin{subfigure}{.5\textwidth}
        \centering
        \includegraphics[width=.8\linewidth]{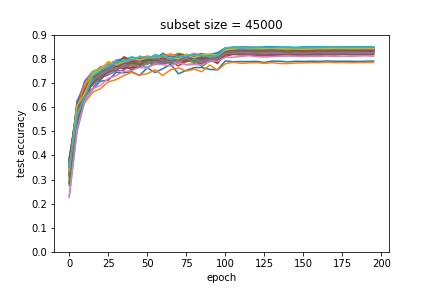}
        \caption{training dataset size = $45000$ images}
        \label{fig:subsetsize45000}
    \end{subfigure}

    \caption{Test accuracies vs. epochs for different number of training dataset sizes for the CIFAR10 experiment.}
    \label{fig:accuracy_vs_epoch_by_subset_size_cifar}
\end{figure}

To get an impression of the models fitted here, there are a number of descriptive visualizations in Figures \ref{fig_full_dataset_epoch_vs_acc_cifar} and \ref{fig_traing_subset_size_vs_test_acc_cifar}. In Figure \ref{fig_traing_subset_size_vs_test_acc_cifar} the test set accuracy for the full training dataset across different epochs and 10 repeats is shown.
In Figure \ref{fig_traing_subset_size_vs_test_acc_cifar} the test accuracy at 195 training epochs for all created training dataset of different size is plotted. As expected, the performance varies more for smaller total training dataset sizes compared to larger training dataset sizes.
Moreover, in Figure \ref{fig:accuracy_vs_epoch_by_subset_size_cifar}, the test accuracy is plotted over training epoch for all training datasets of 4 different training dataset sizes.
Overall, the descriptive analysis did not yield any concerns in terms of outliers or unexpected behavior of the data.

\subsection{Model fitting results}

\begin{figure}
    \begin{subfigure}{.33\textwidth}
        \centering
        \includegraphics[width=.8\linewidth]{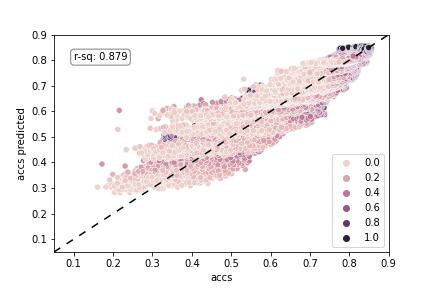}
        \caption{Train dataset.}
        \label{fig:powerlaw_acc_classes_linear_epoch_linear_train}
    \end{subfigure}%
    \begin{subfigure}{.33\textwidth}
        \centering
        \includegraphics[width=.8\linewidth]{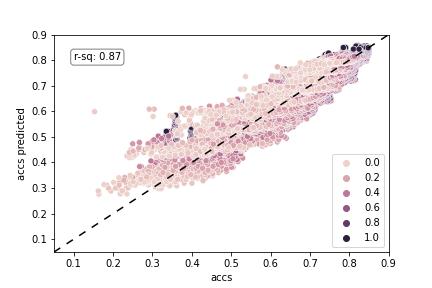}
        \caption{Test dataset.}
        \label{fig:powerlaw_acc_classes_linear_epoch_linear_val}
    \end{subfigure}
    \begin{subfigure}{.33\textwidth}
        \centering
        \includegraphics[width=.8\linewidth]{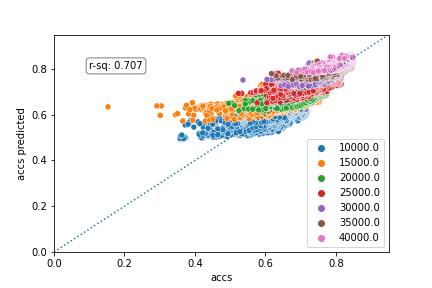}
        \caption{Forward testing on test dataset.}
        \label{fig:powerlaw_acc_classes_linear_epoch_linear_forward_val}
    \end{subfigure}
    \caption{Prediction plots for powerlaw function 0 on the training dataset (left), test dataset (middle) and using forward testing on the test dataset (right) for the CIFAR10 experiment.}
    \label{fig:powerlaw_prediction_plot_fct_0}
\end{figure}

\begin{figure}
    \begin{subfigure}{.33\textwidth}
        \centering
        \includegraphics[width=.8\linewidth]{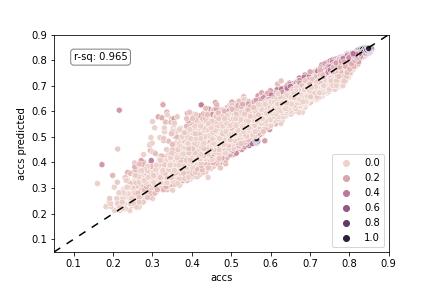}
        \caption{Train dataset.}
        \label{fig:powerlaw_acc_classes_arctan_epoch_artan_train}
    \end{subfigure}%
    \begin{subfigure}{.33\textwidth}
        \centering
        \includegraphics[width=.8\linewidth]{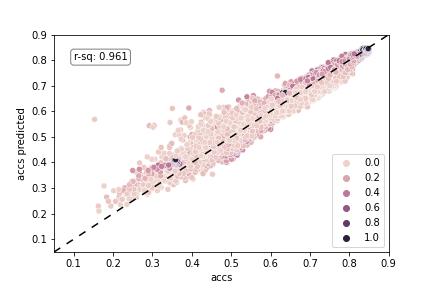}
        \caption{Test dataset.}
        \label{fig:powerlaw_acc_classes_arctan_epoch_artan_val}
    \end{subfigure}
    \begin{subfigure}{.33\textwidth}
        \centering
        \includegraphics[width=.8\linewidth]{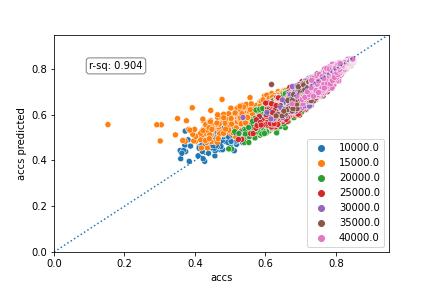}
        \caption{Forward testing on test dataset.}
        \label{fig:powerlaw_acc_classes_arctan_epoch_artan_forward_val}
    \end{subfigure}
    \caption{Prediction plots for powerlaw function 1 on the training dataset (left), test dataset (middle) and using forward testing on the test dataset (right) for the CIFAR10 experiment.}
    \label{fig:powerlaw_prediction_plot_fct_1}
\end{figure}

\begin{figure}
    \begin{subfigure}{.33\textwidth}
        \centering
        \includegraphics[width=.8\linewidth]{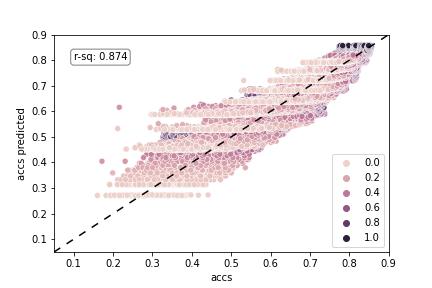}
        \caption{Train dataset.}
        \label{fig:powerlaw_acc_total_n_epoch_train}
    \end{subfigure}%
    \begin{subfigure}{.33\textwidth}
        \centering
        \includegraphics[width=.8\linewidth]{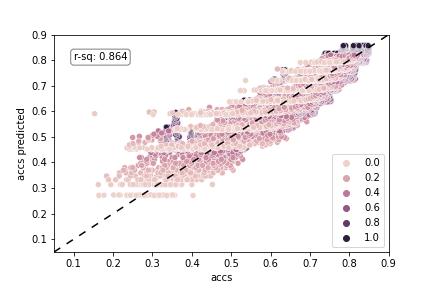}
        \caption{Test dataset.}
        \label{fig:powerlaw_acc_total_n_epoch_val}
    \end{subfigure}
    \begin{subfigure}{.33\textwidth}
        \centering
        \includegraphics[width=.8\linewidth]{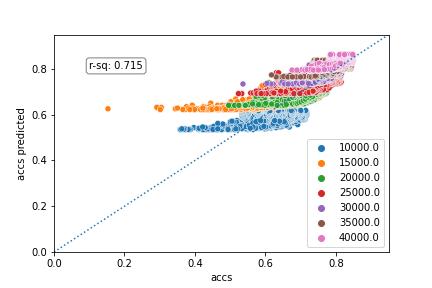}
        \caption{Forward testing on test dataset.}
        \label{fig:powerlaw_acc_total_n_epoch_forward_val}
    \end{subfigure}
    \caption{Prediction plots for powerlaw function 2 on the training dataset (left), test dataset (middle) and using forward testing on the test dataset (right).}
    \label{fig:powerlaw_prediction_plot_fct_2}
\end{figure}

\begin{figure}
    \begin{subfigure}{.33\textwidth}
        \centering
        \includegraphics[width=.8\linewidth]{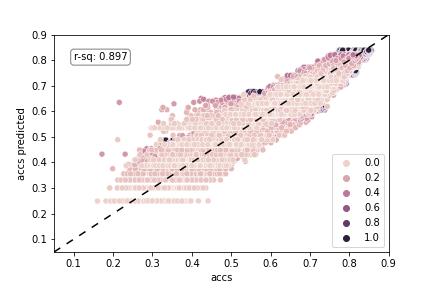}
        \caption{Train dataset.}
        \label{fig:powerlaw_acc_total_n_arctan_epoch_arctan_train}
    \end{subfigure}%
    \begin{subfigure}{.33\textwidth}
        \centering
        \includegraphics[width=.8\linewidth]{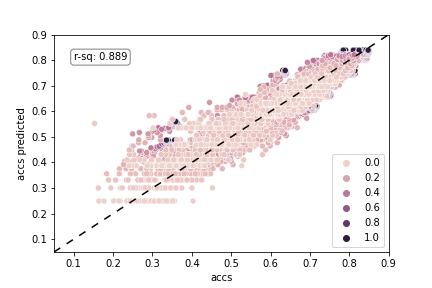}
        \caption{Test dataset.}
        \label{fig:powerlaw_acc_total_n_arctan_epoch_arctan_val}
    \end{subfigure}
    \begin{subfigure}{.33\textwidth}
        \centering
        \includegraphics[width=.8\linewidth]{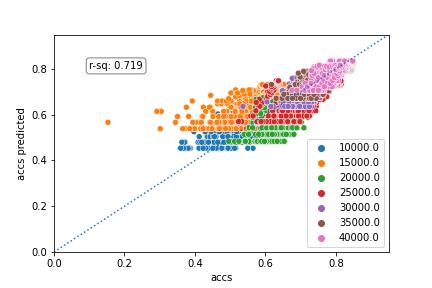}
        \caption{Forward testing on test dataset.}
        \label{fig:powerlaw_acc_total_n_arctan_epoch_arctan_forward_val}
    \end{subfigure}
    \caption{Prediction plots for powerlaw function 3 on the training dataset (left), test dataset (middle) and using forward testing on the test dataset (right).}
    \label{fig:powerlaw_prediction_plot_fct_3}
\end{figure}

This section shows the results of the different model fittings for the CIFAR10 experiment. In preparation for model fitting, the data have been transformed in the following way:
\begin{itemize}
    \item All model results for epochs $< 10$ have been removed from the dataset. Mainly because these data are highly variable.
    \item From the individual class counts an overall training dataset size is calculated per dataset.
    \item All class counts (also the total dataset size) have been scaled to be between $[0, 1]$, as well as the epochs.
    \item As the test accuracies are already between $[0,1]$ these have not been further standardized.
\end{itemize}
Following these descriptive plots, a number of different models are fitted, all using the above described dataset.
In Table \ref{table:cifar_model_overview} an overview of the fitted models is given.
Model $(0)$ is easily motivated by the idea of generalized linear models: the overall dataset size, as used in many other papers cited here, is replaced by a linear combination of the individual class counts and the epoch, which introduces a larger number of parameters and the capability to account for different importances of the classes. 
This is in equivalence to many methods in generalized linear models as can be read in the standard reference \cite{GeneralizedLinearModels}.
However when looking at the modeling results for this model in Figure (\ref{fig:powerlaw_prediction_plot_fct_3}), the prediction plots still look like some signal in the dataset is not captured by the fitted powerlaw model.
One obvious next step would have been to increase the polynomial degree of the linear combination, i.e. to add interactions and/or quadratic terms. However, this did not deliver a substantial better fit.
Hence the 2-parameter arctan effect is introduced. It is motivated by the need to have 2 different types of influences of an input: 
A class count can have a very strong or weak overall maximum effect to the response fitted, and an input can actually achieve it's maximum effect with either very small or only very high counts.
This is achieved by the 2-parameter arctan effect $\beta_{., 1} \arctan{(\beta_{., 2} x)}$. The $\beta_{.,1}$ parameter describes the maximum possible influence of an input 
and the $\beta_{., 2}$ describes the necessary size/counts of input $x$ to achieve this maximum influence.
One advantage of the model over just adding higher degree polynomial terms is also that the parameters have a clear interpretation.
As a reference to existing literature a standard powerlaw model $(2)$ has been fitted, using both the total training dataset size and the number of epochs.
In order to have a fair comparison, model $(3)$ is fitted, which is using the same 2-parameter arctan effect as in model $(1)$ for both the total training dataset size as well as for the number of training epochs.
All of the these models are fitted using the \verb|curve_fit| function from the python package \verb|scipy.optimize|.
The resulting parameter estimates are shown in Tables \ref{table:cifar10_param_values_model_0} to \ref{table:cifar10_param_values_model_3}.
For each model, there is a corresponding triple of prediction plots (one for the train data set and one for the validation dataset) in Figures \ref{fig:powerlaw_prediction_plot_fct_0} to \ref{fig:powerlaw_prediction_plot_fct_3}.
Looking at Table \ref{table:cifar_model_overview}, the model using the individual class counts and the epoch with a 2 parameter arctan model performs best. 
Also, for this model, the prediction plots look most reasonable, i.e. scattering rather randomly close to the diagonal.

\begin{table}[h!]
    \centering
    \begin{tabular}{|r|c|c|c|c|c|c|}
        \hline
        short description   & no  & linear combination                               & train loss & test loss & train acc $r^2$ & test acc $r^2$ \\
        \hline
        full linear model          & $0$ & $\beta_e x_{epoch} + \sum_{i = 1}^c \beta_c n_c$ & 0.0021    & 0.0023   & $87.9\%$        & $87.0\%$       \\
        \hline
        full arctan model & $1$& \makecell{$\beta_{e, 1} \arctan{(\beta_{e, 2} x_{epoch})}$ \\  $+ \sum_{i = 1}^c \beta_{i, 1} \arctan{(\beta_{i, 2} x_{i})}$}& 0.0006 & 0.0007 & $96.5\%$ & $96.1\%$ \\
        \hline
        $total_n$ linear model & $2$ & $\beta_e x_{epoch} + \beta_n n_{total}$          & 0.0022    & 0.0024   & $87.4\%$        & $86.4\%$       \\
        \hline
        $total_n$ arctan model &$3$&\makecell{ $\beta_{e, 1} \arctan{(\beta_{e, 2} x_{epoch})}$ \\ $+ \beta_{n, 1} \arctan{(\beta_{n, 2} n_{total})}$} & 0.0018 & 0.0019 & $89.7\%$ & $88.9\%$ \\
        \hline
    \end{tabular}
    \caption{Powerlaw model overview for CIFAR10 experiment.}
    \label{table:cifar_model_overview}
\end{table}

\begin{table}[h!]
    \centering
    \begin{tabular}{|c|c|c|c|c|c|c|c|c|c|c|c|c|c|}
        \hline
        a     & b    & c    & plane & car  & bird & cat  & deer & dog  & frog & horse & ship & truck & epoch \\
        \hline
        -0.41 & 0.23 & 0.65 &   1.21 & 1.63 &  1.40 & 1.57 &  2.11 & 1.10 &  1.81 &   1.95 &  1.56 &   1.94 &   3.54  \\
        \hline
    \end{tabular}
    \caption{Estimated parameters for model (0).}
    \label{table:cifar10_param_values_model_0}
\end{table}

\begin{table}[h!]
    \centering
\begin{tabular}{|l|c|c|c|c|c|c|c|c|c|c|c|c|c|c|c|}
\hline
     &     a &    b  &    c &  plane     &    car    &    bird   &  cat   &   deer     &  dog    &  frog     &  horse    &  ship &  truck    &  epoch  \\
\hline
$\beta_{.,1}$    & -0.01 & 0.94  & 0.10 &     0.52   &     0.60  &     0.49  &  0.40  &    0.57    &   0.45  &    0.62   &     0.59  &  0.67 &     0.62  &  1.25   \\
\hline
$\beta_{.,2}$     & -     & -     &      &     12.78  &   15.32   & 5.97      & 4.59   & 6.05       & 9.63    & 19.13     &   9.93    & 29.12 & 21.25     &  6.31  \\
\hline
\end{tabular}
\caption{Estimated parameters for model (1).}
\label{table:cifar10_param_values_model_1}
\end{table}

\begin{table}[h!]
    \centering
    \begin{tabular}{|c|c|c|c|c|}
        \hline
        a    & b    & c    & total\_n & epoch \\
        \hline
        0.27 & 0.50 & 0.53 &     0.96 &   0.26  \\
        \hline
    \end{tabular}
    \caption{Estimated parameters for model (2).}
    \label{table:cifar10_param_values_model_2}
\end{table}

\begin{table}[h!]
    \centering
    \begin{tabular}{|l|c|c|c|c|c|}
        \hline
            &    a  &    b  &    c &  total\_n &   epoch \\
        \hline
        $\beta_{.,1}$  & 0.25  & 0.70  & 0.62 &      0.75 &   0.19   \\
        \hline
        $\beta_{.,2}$   & -     & - & -  & 1.27      & 5.53 \\
        \hline
    \end{tabular}
    \caption{Estimated parameters for model (3).}
    \label{table:cifar10_param_values_model_3}
\end{table}

Looking at the results in Table (\ref{table:cifar_model_overview}), there is a clear winner in terms of model performance: model $(1)$ performs best.
The additional cost of having more parameters to optimize seems to have paid off, the $r^2$ of model $(2)$ is about 8 to 9 \% better then the other 3 models.
Also the prediction plots look more randomly scattered around the diagonal.
Looking at the parameters for model $(1)$ in Table (\ref{table:cifar10_param_values_model_1}), all $\beta$ parameters have positive values (which also holds true for the other models), which was not enforced as an unconstrained optimization has been used.
For model $(1)$ the classes truck and ship both yielded the highest $\beta_{.,1}$ and $\beta_{.,2}$ value, indicating that these classes would be most meaningful to overweight during additional labeling efforts.

\section{EMNIST}

\begin{figure}
    \begin{subfigure}{.5\textwidth}
        \centering
        \includegraphics[width=.8\linewidth]{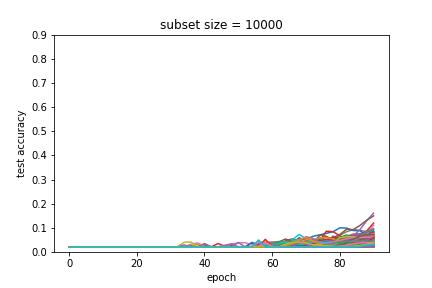}
        \caption{training dataset size = $10000$ images}
        \label{fig:subsetsize5000}
    \end{subfigure}%
    \begin{subfigure}{.5\textwidth}
        \centering
        \includegraphics[width=.8\linewidth]{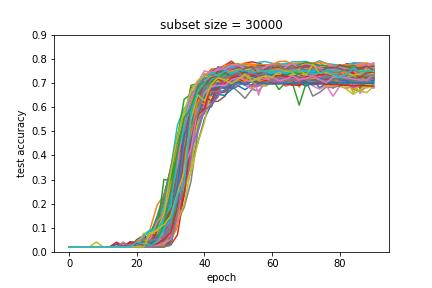}
        \caption{training dataset size = $30000$ images}
        \label{fig:subsetsize20000}
    \end{subfigure}
    \begin{subfigure}{.5\textwidth}
        \centering
        \includegraphics[width=.8\linewidth]{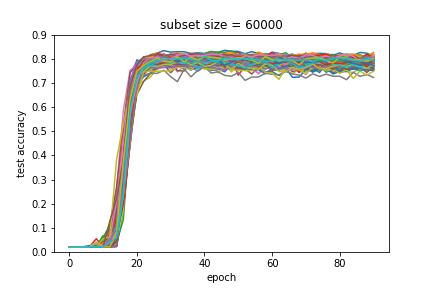}
        \caption{training dataset size = $60000$ images}
        \label{fig:subsetsize30000}
    \end{subfigure}%
    \begin{subfigure}{.5\textwidth}
        \centering
        \includegraphics[width=.8\linewidth]{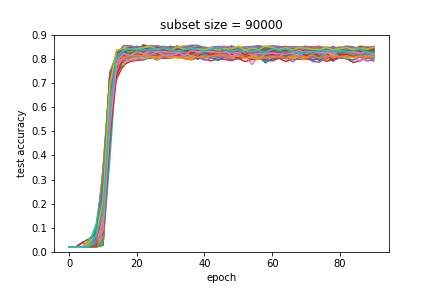}
        \caption{training dataset size = $90000$ images}
        \label{fig:subsetsize45000}
    \end{subfigure}

    \caption{Test accuracies vs. epochs for different number of training dataset sizes for the EMNIST dataset.}
    \label{fig:emnist_accuracy_vs_epoch_by_subset_size}
\end{figure}

\begin{figure}
    \begin{subfigure}{.5\textwidth}
        \centering
        \includegraphics[width=.8\linewidth]{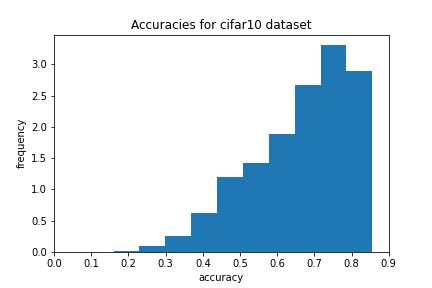}
        \label{fig:accuracy_distribution_cifar10}
    \end{subfigure}%
    \begin{subfigure}{.5\textwidth}
        \centering
        \includegraphics[width=.8\linewidth]{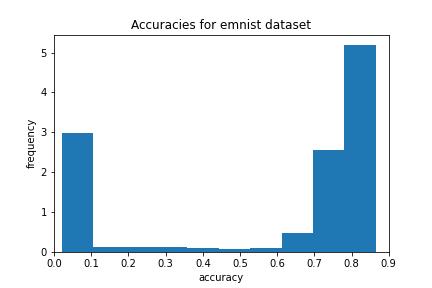}
        \label{fig:accuracy_distribution_emnist}
    \end{subfigure}
    \caption{Accuracy distributions for models fitted to CIFAR10 and EMNIST dataset.}
    \label{fig:acc_distributions}
\end{figure}

The EMNIST dataset is an extension of the famous MNIST dataset, i.e. has more classes. The version used here is the balanced EMNIST dataset, consisting of 47 different characters, each class having 2400 images in the train dataset and 400 images in a validation dataset \cite{cohen2017emnist}.  
Main motivation for applying the suggested method on this dataset as well is that it contains substantially more label classes, which is often more realistic for real world applications compared to just 10 classes.
The model architecture used here for the image classification is the \verb|mobilenet_v3_large| model from the pytorch model zoo \cite{howard2019mobilenetv3}.
As for the CIFAR10 dataset, the suggested algorithm has been applied with the following details:
\begin{itemize}
    \item Subset sizes $[10000, 20000, \dots, 100000]$, per subset size 140 different settings for the train dataset and 70 different settings for the validation dataset.
    \item For each setting, the \verb|mobilenet_v3_large| model has been fitted, using the SGD optimizer (with learning rate $=0.1$, momentum $=0.9$, weight decay $=1e-4$ and a cross entropy loss function), a batch size of 2048 for 90 epochs. 
\end{itemize}

Here, a slightly different modeling approach compared to the CIFAR10 dataset is chosen. 
First, the resulting accuracies are differently distributed (see Figure (\ref{fig:acc_distributions})), which motivates to not use a powerlaw, but an arctan model as this allows better for modelling a bimodal dataset.
Also, directly using all classes as features for the model is possible but not necessarily recommendable due to the high number of classes.
Hence, an interactive modeling approach has been chosen here, specifically a forward selection approach using a residual sum of squares as selection criterion. 
$$y = a + c* \arctan{\left(\sum_{j = 1}^{J}{\theta_j x_j} + b \right)}$$
The selection of which features are included is determined by forward selection. In short, in forward selection, an empty model is initialized, and iteratively new features are added to the model which are minimising the residual sum of squares, given the already chosen features from previous steps.
An important aspect in this method is the set of candidate features. Here the following set of features is used: all class labels, total training dataset size, epoch and all 2 factor interactions of these features, which results here in a total of 1225 candidate features.
These candidates features have been added until the reduction of residual sum of square between two runs of feature selections is below 0.1.
This is to a large degree arbitrary and could be changed to different values.  
The resulting model has 10 features (see Table \ref{table:param_values_emnist}) which resulted in a model with an $r^2 = 0.987$. While there are some features which are expectedly added (total dataset size, epoch and their interaction), there are also more interesting features selected.
For example, the interaction between the characters $7$ and $J$ is automatically selected which intuitively makes sense, as these characters are very similar. As the coefficient of this interaction is positive, this indicates that it is beneficial for a higher model performance to have both a high number of images with $7$ and $J$.

\begin{table}[h!]
    \centering
\begin{tabular}{r|c|c}

    parameter name &  parameter value &    residual sum of squares \\
    \hline
    a (arctan)                                  &         0.39 &    -    \\
    b (arctan)                                  &        18.25 &    -    \\
    c (arctan)                                  &         0.27 &    -    \\
    total\_training\_size                &        16.29 & 943.86 \\
    epochs\_trained                     &        15.29 & 517.94 \\
    epochs\_trained:total\_training\_size &         9.06 &  23.40 \\
    Y:total\_training\_size              &         0.17 &  23.00 \\
    W:total\_training\_size              &         0.15 &  22.74 \\
    q                                  &         0.09 &  22.57 \\
    B:h                                &         0.08 &  22.43 \\
    seven:J                            &         0.08 &  22.31 \\
    four:total\_training\_size           &         0.11 &  22.19 \\
    six:O                              &        -0.06 &  22.11 \\
    \end{tabular}
\caption{Estimated parameters for forward selected model for the EMNIST dataset.}
\label{table:param_values_emnist}
\end{table}

\begin{figure}
    \begin{subfigure}{.33\textwidth}
        \centering
        \includegraphics[width=.98\linewidth]{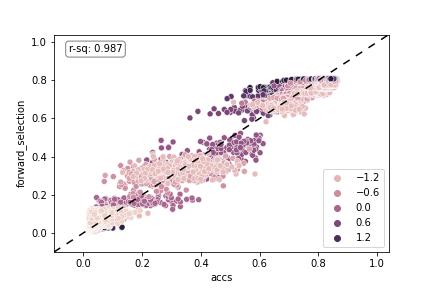}
        \caption{Prediction plot for the EMNIST training dataset.}
        \label{fig:emnist_prediction_plot_train}
    \end{subfigure}%
    \begin{subfigure}{.33\textwidth}
        \centering
        \includegraphics[width=.98\linewidth]{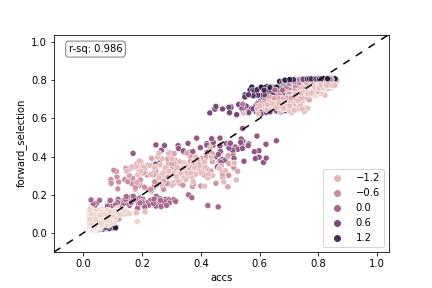}
        \caption{Prediction plot for the EMNIST test dataset.}
        \label{fig:emnist_prediction_plot_val}
    \end{subfigure}
    \begin{subfigure}{.33\textwidth}
        \centering
        \includegraphics[width=.98\linewidth]{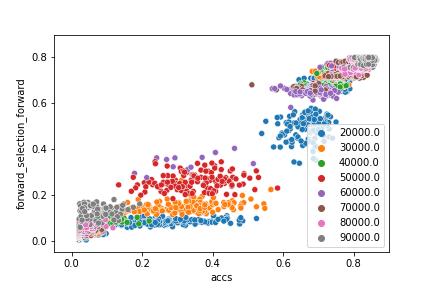}
        \caption{Prediction plot for the forward prediction on the test dataset.}
        \label{fig:emnist_prediction_plot_val}
    \end{subfigure}
    \caption{Prediction plots for arctan forward selection model for EMNIST dataset.}
    \label{fig:emnist_prediction_plot}
\end{figure}

\section{Discussion summary}

The methods presented in this paper do not just predict model performance based on overall training dataset size, but specific to available training data per label class.
This allows for a more accurate prediction of model performance and it also can be more informative.
However, there are also downsides: The number of trained models needs to be far higher in order to fit more detailed models.
Hence this approach is not necessarily meaningful for situations where only a limited training budget is available but rather for situations where the cost of labeling is high compared to the cost of training models.
Further improvements of this method could be to modify it for unbalanced training datasets and also to modify it for other tasks than classification.

\pagebreak
\bibliographystyle{ieeetr}
\bibliography{lit}

\section*{Appendix}

\begin{table}[h!]
    \centering
    \begin{tabular}{lrrrrrrrrrrrrr}
        row & accs & plane & car & bird & cat & deer & dog & frog & horse & ship & truck & epochs & $total_n$ \\
        \hline
        0   & 0.18 & 145   & 31  & 97   & 496 & 1096 & 307 & 2382 & 10    & 373  & 63    & 10     & 5000      \\
        1   & 0.21 & 145   & 31  & 97   & 496 & 1096 & 307 & 2382 & 10    & 373  & 63    & 15     & 5000      \\
        2   & 0.20 & 145   & 31  & 97   & 496 & 1096 & 307 & 2382 & 10    & 373  & 63    & 20     & 5000      \\
        3   & 0.25 & 145   & 31  & 97   & 496 & 1096 & 307 & 2382 & 10    & 373  & 63    & 25     & 5000      \\
        4   & 0.24 & 145   & 31  & 97   & 496 & 1096 & 307 & 2382 & 10    & 373  & 63    & 30     & 5000      \\
    \end{tabular}
    \caption{First few lines of the underlying data table used for fitting the powerlaw models.}
    \label{table:datatable_results}
\end{table}

\end{document}